\documentclass[conference]{IEEEtran}
\IEEEoverridecommandlockouts
\usepackage{cite}
\usepackage{amsmath,amssymb,amsfonts}
\usepackage{algorithmic}
\usepackage{graphicx}
\usepackage{textcomp}
\usepackage{xcolor}
\def\BibTeX{{\rm B\kern-.05em{\sc i\kern-.025em b}\kern-.08em
    T\kern-.1667em\lower.7ex\hbox{E}\kern-.125emX}}

\usepackage{times}
\usepackage{epsfig}
\usepackage{graphicx}
\usepackage{amsmath}
\usepackage{amssymb}

\usepackage{subfigure}
\usepackage{multirow}
\usepackage{diagbox}
\usepackage{booktabs}
\usepackage{arydshln}
\usepackage[normalem]{ulem}
\useunder{\uline}{\ul}{}
\usepackage[misc]{ifsym}

\usepackage{amsthm, amsmath , amssymb}
\usepackage{mathrsfs}

\let\OLDthebibliography\thebibliography
\renewcommand\thebibliography[1]{
  \OLDthebibliography{#1}
  \setlength{\parskip}{0pt}
  \setlength{\itemsep}{0pt plus 0.3ex}
}

\begin{document}

\title{Learning Disentangled Representation with Mutual Information Maximization for Real-Time UAV Tracking\\
\thanks{* These authors contributed equally.}
}

\author{\IEEEauthorblockN{1\textsuperscript{st} Xucheng Wang*}
\IEEEauthorblockA{\textit{Guilin University of Technology} \\
Guilin, China \\
xcwang@glut.edu.cn}\\
\IEEEauthorblockN{3\textsuperscript{rd} Hengzhou Ye}
\IEEEauthorblockA{\textit{Guilin University of Technology} \\
Guilin, China \\
2002018@glut.edu.cn}
\and
\IEEEauthorblockN{2\textsuperscript{nd} Xiangyang Yang*}
\IEEEauthorblockA{\textit{Guilin University of Technology} \\
Guilin, China \\
xyyang317@163.com}\\
\IEEEauthorblockN{4\textsuperscript{th} Shuiwang Li {\textrm\Letter}}
\IEEEauthorblockA{\textit{Guilin University of Technology} \\
Guilin, China \\
lishuiwang0721@163.com}
}

\maketitle

\begin{abstract}
Efficiency has been a critical problem in UAV tracking due to limitations in computation resources, battery capacity, and unmanned aerial vehicle maximum load. Although discriminative correlation filters (DCF)-based trackers prevail in this field for their favorable efficiency, some recently proposed lightweight deep learning (DL)-based trackers using model compression demonstrated quite remarkable CPU efficiency as well as precision. Unfortunately, the model compression methods utilized by these works, though simple, are still unable to achieve satisfying tracking precision with higher compression rates. This paper aims to exploit disentangled representation learning with mutual information maximization (DR-MIM) to further improve DL-based trackers' precision and efficiency for UAV tracking. The proposed disentangled representation separates the feature into an identity-related and an identity-unrelated features. Only the latter is used, which enhances the effectiveness of the feature representation for subsequent classification and regression tasks. Extensive experiments on four UAV benchmarks, including UAV123@10fps, DTB70, UAVDT and VisDrone2018, show that our DR-MIM tracker significantly outperforms state-of-the-art UAV tracking methods.
\end{abstract}

\begin{IEEEkeywords}
UAV tracking, Disentangled representation, Mutual information
\end{IEEEkeywords}

\section{Introduction}
With the recent development and deployment of unmanned aerial vehicles (UAVs) in numerous applications, UAV tracking has arisen as a new challenge and has sparked increased interest in visual tracking \cite{li2020autotrack,cao2021hift,Wang2022RankBasedFP,Wu2022FisherPF,articlewxc}. However, UAV tracking faces more difficult challenges, compared with general scenarios. Motion blur, severe occlusion, extreme viewing angle, and scale changes, on the one hand, have greatly challenged the precision of the UAV tracking algorithms; on the other hand, limited computing resources, low power consumption requirements, battery capacity limitations, and UAV's maximum load pose a big challenge  to its efficiency  \cite{Wang2022RankBasedFP,Wu2022FisherPF,li2021learning}.


As efficiency is a bottleneck in UAV tracking at the current technological level, discriminative correlation filters (DCF)-based trackers prevail in this field for their favorable efficiency on a single CPU. Although tracking precisions of DCF-based trackers have substantially increased, they are still unable to match the majority of cutting-edge deep learning-based trackers \cite{li2020autotrack, huang2019learning,Zhang2022TrackingSA,Zhang2023TsfmoAB}. In order to break through the restricted precision of DCF-based trackers, some lightweight DL-based trackers were proposed very recently for UAV tracking \cite{Wang2022RankBasedFP,Wu2022FisherPF,Liu2022GlobalFP,Zhong2022EfficiencyAP}, which focused on demonstrating model compressing techniques such as filter pruning could be exploited to obtain lightweight DL-based trackers that remarkably boost both precision and efficiency for UAV tracking. Unfortunately, simple as the filter pruning methods utilized by these works are, the tracking precision and efficiency so achieved are still very limited and far from satisfying.   
In this paper, we aim to learn disentangled representation based on mutual information maximization to further improve both DL-based trackers' precision and efficiency for UAV tracking.

Disentangled representation is a type of distributed feature representation that enhances model performance by uncovering and disentangling latent explanatory factors in observed data. In a disentangled representation, information on an individual factor value is easily accessed and is robust to changes in the input that do not affect this factor.
Learning to solve a downstream problem using a disentangled representation has demonstrated tremendous success in numerous tasks in recent years, including image classification, voice transfer, point clouds processing, and etc \cite{Zhang2020FaceAV, Yuan2021ImprovingZV}. Very recently, mutual information (MI) estimation has been exploited to aid disentangled representation learning, which has proven very successful in applications such as face recognition, human pose estimation, and image retrieval \cite{Zhao2021LearningVH, Guo2019LearningDR}. Although MI is known to be hard to estimate and its success might depend on the inductive bias in both the choice of feature extractor architectures and the parametrization of the employed MI estimators \cite{tschannen2019mutual}, we believe that exploiting MI to learn disentangled representation is a generic method that can be applied to visual tracking as well. In this paper, we attempt to explore learning disentangled representation via mutual information maximization (MIM) to achieve better and more compact feature representation in order to improve both precision and efficiency of lightweight DL-based trackers for UAV tracking, which hasn't been well explored to the best of our knowledge. We summarize our contributions as follows:

\begin{itemize}
	\item We make the first attempt to explore learning disentangled representation via mutual information maximization for UAV tracking in order to obtain lightweight DL-based trackers of better tracking precision and efficiency.
	\item We propose the DR-MIM tracker that learns more effective and more compact disentangled representations for UAV tracking, achieving remarkable tracking efficiency and precision. 
	\item We demonstrate the proposed method on four public UAV benchmarks. Experimental results show that the proposed DR-MIM tracker achieves state-of-the-art performance.
\end{itemize}

\section{Related Works}

\subsection{UAV Tracking Methods}

Modern trackers can be roughly divided into two classes: DCF-based trackers and DL-based ones. The former prevail in UAV tracking for their more favorable efficiency. Despite their relatively higher efficiency, they hardly maintain robustness under challenging conditions because of the poor representation ability of handcrafted features \cite{li2021learning,Li2021LearningRC,li2020autotrack,huang2019learning,li2020asymmetric,Li2021EquivalenceOC}. 
To substantially improve tracking precision and robustness, some DL-based trackers have been developed for UAV tracking recently. For instance, Cao et al. \cite{cao2021hift} proposed a hierarchical feature transformer to achieve interactive fusion of spatial (shallow layers) and semantics cues (deep layers) for UAV tracking. 
Huang et al. \cite{Cao2022TCTrackTC} presented a comprehensive framework to fully exploit temporal contexts with a proposed adaptive temporal transformer for aerial tracking. However, the efficiency of these methods is still much lower than most DCF-based trackers. To further improve efficiency of DL-based trackers for UAV tracking, model compression techniques have been recently utilized to reduce model size and thus to improve efficiency \cite{Wang2022RankBasedFP,Wu2022FisherPF}. Unfortunately, the model compression methods utilized by these works, though simple, are still unable to achieve satisfying tracking precision with higher compression rates. To further improve DL-based trackers’ precision and efficiency for UAV tracking, in this paper we exploit disentangled representation learning with mutual information maximization to obtain more effective and more compact representation.

\subsection{Disentangled Representation Learning with Mutual Information Estimation}
Disentangled representation is an unsupervised learning method that divides each feature into precisely specified variables and encodes them as distinct dimensions, which are semantically meaningful. In recent years, learning to solve a down-stream task from a disentangled representation has shown great success in many tasks. For instance, Zhang et al. \cite{Zhang2020FaceAV} used disentangled representation learning to disentangles an image into a liveness feature and a content feature, and the former was further used for classification. Yuan et al. \cite{Yuan2021ImprovingZV} proposed a zero-shot voice transfer method via disentangled representation learning, in which each input voice is disentangled into two separated low-dimensional features in the embedding space, encoding speaker-related style and voice content, respectively.

Recently it has witnessed a revival of the InfoMax principle \cite{Linsker1988SelforganizationIA} that learns a representation which maximizes/minimizes the mutual information (MI) between the input and its representation, possibly subject to some structural constraints, where MI is a measure of the mutual dependence between the two random variables, which, more specifically, quantifies the "amount of information" obtained about one random variable by observing the other random variable. This principle has been applied to aid disentangled representation learning in various applications recently, which turned out to be very helpful. For instance, 
Zhao et al. \cite{Zhao2021LearningVH} proposed to train a network using cross-view mutual information maximization to disentangle pose-dependent as well as view-dependent factors from 2D human poses, which maximized MI of the same pose performed from different viewpoints in a contrastive learning manner.
Li et al. \cite{Li2022TowardsLD} presented a framework for multi-level disentanglement for time series data by covering both individual latent factors and group semantic segments by introducing a MI maximization term between the observation space to the latent space to alleviate the Kullback-Leibler (KL) vanishing problem while preserving the disentanglement property. However, to our knowledge, learning disentangled representation via MI estimation hasn't been utilized to UAV tracking, which we believe could be very helpful as well since the InfoMax principle is very general. In this work, we make the first attempt and demonstrate its usefulness to UAV tracking.

\section{Learning Disentangled Representation with Mutual Information Maximization}
\begin{figure*}[h]
	\centering
	\includegraphics[width=0.95\textwidth,height=0.32\textwidth]{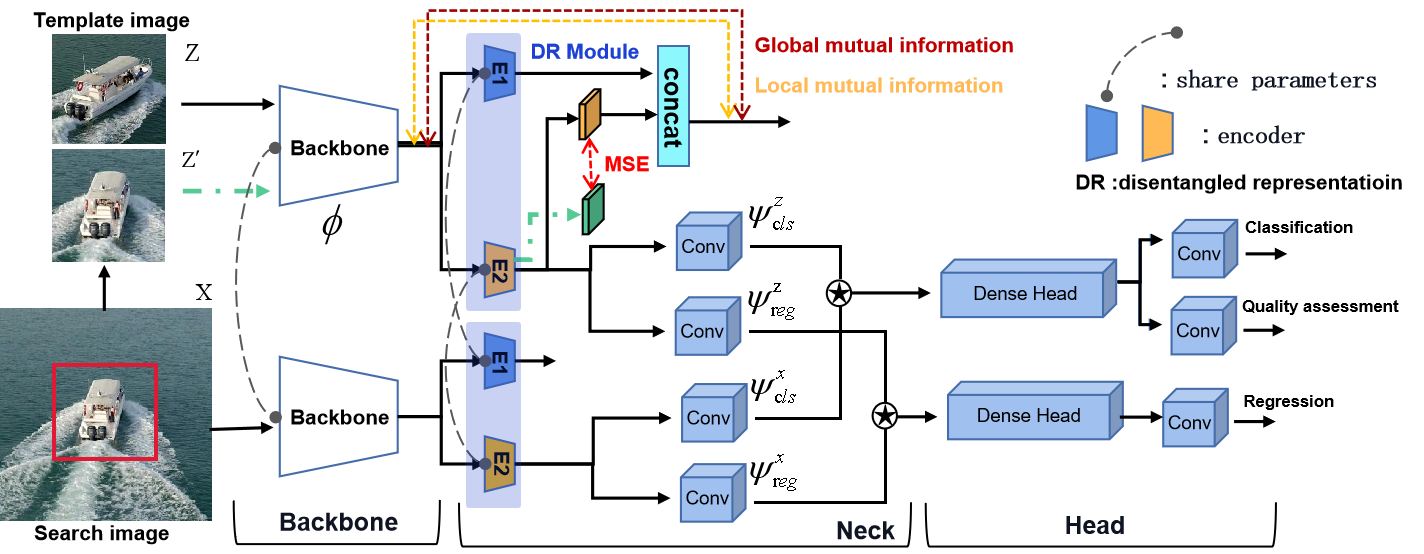}
	\caption{An illustration of the proposed DR-MIM method. Note that $\psi_{cls}^{\cdot}$ and $\psi_{reg}^{\cdot}$ denote the task-specific convolutional layers for classification and regression, respectively. } \label{P-saimfc++_overview}
\end{figure*}

\subsection{DR-MIM Overview}
The proposed DR-MIM consists of a backbone, a neck and a head network. As illustrated in Fig. \ref{P-saimfc++_overview}, the backbone consists of a Siamese network, which has a template branch and a search branch. The template branch and the search branch share the same backbone network $\phi(\cdot)$ and take the template image $Z$ and the search image $X$ as input. The neck contains two shared-architecture disentangled representation (DR) modules and four convolutional layers used to adjust feature sizes. The DR modules disentangle the backbone features into identity-related and identity-unrelated features using two separate encoders, namely $E_1$ and $E_2$. 
In training phase, the disentangled features will be concatenated to compute its mutual information with the backbone features. The identity-related features of an input template $Z$ and that of the template $Z'$ extracted from the search image are used to construct a similarity loss that enforces the same target to have close identity-related feature representation at different frames. The head consists of two dense head branches followed by three convolutional layers to produce outputs for classification, quality assessment, and regression tasks, respectively.
Note that, in order to improve the effectiveness of the feature representation, only the identity-related features are used for subsequent classification and regression tasks. Specifically, the identity-related features from the template branch and the search branch are adjusted in the neck and then coupled with cross-correlation before they are finally fed into the classification and regression heads.
The coupled features are formulated by:
\begin{equation}
\small
    f_l(Z,X) = E_2(\psi_l^z(\phi(Z))) \star E_2(\psi_l^x(\phi(X))), l\in \{cls, reg\},
\end{equation}
where $\star$ denotes the cross-correlation operation, $\psi_{cls}^x(\cdot)$ and $\psi_{reg}^x(\cdot)$ denote the task-specific layer for classification and regression, respectively, whose outputs are of the same size. $\psi_{cls}^z(\cdot)$ and $\psi_{reg}^z(\cdot)$ are interpreted likewise.
All these architectures are inherited from P-SiamFC++ \cite{Wang2022RankBasedFP} but with a different pruning ratio and a newly-proposed disentangled representation module aiming at obtaining more compact and more effective feature representation, which will be detailed in the following subsection. It is worthy of note that the layer-wise pruning ratios used in P-SiamFC++ \cite{Wang2022RankBasedFP} is laborious and time-consuming to determine, we use a global pruning ratio of 0.5 instead in this paper.

\subsection{Disentangled Representation via Mutual Information Maximization}
To start with, we summarize the concept of mutual information (MI) and introduce related notations. Let $x\in \mathscr{X}$ and $y\in \mathscr{Y}$ be two random variables. The MI between $x$ and $y$, denoted by $I(x,y)$, formally can be expressed as follows:
\begin{equation}\label{Eq_MI}
\small
\begin{split}
\mathbb{E}_{p(x,y)}\left [ log\frac{p(x,y)}{p(x)p(y)}\right ]=D_{KL}(p(x,y)||p(x)p(y)),
\end{split}
\end{equation}
where $p(x,y)$ denotes the joint probability distribution, $p(x)$ and $p(y)$ are the marginals, $D_{KL}$ is the Kullback–Leibler divergence (KLD). In practice, it is difficult to estimate MI as typically samples are all that we have access to but not the underlying distributions \cite{Poole2019OnVB}. 
In this work, we utilize the MI estimator Deep InfoMax \cite{Hjelm2019LearningDR}, which is based on Jensen-Shannon divergence (JSD) instead of KLD, to learn disentangled representation for UAV tracking. 
This Jensen-Shannon MI estimator, denoted by $\hat{I}_{\theta}^{(JSD)}(x,y)$, is defined as follows:
\begin{equation}\label{Eq_JSDMI}
\small
\begin{split}
\mathbb{E}_{p(x,y)}\left [ -\alpha(-T_{\theta}(x,y))\right ]-\mathbb{E}_{p(x)p(y)}\left [ \alpha(T_{\theta}(x,y))\right ],
\end{split}
\end{equation}

where $T_{\theta}:\mathscr{X}\times \mathscr{Y}\rightarrow \mathbb{R}$ is a neural network parameterized by $\theta$, $\alpha(z)=log(1+e^z)$ is the softplus function. Hjelm et al. \cite{Hjelm2019LearningDR} proposed to maximize a global and a local MI between an input image $x$ and its feature representation $y=\texttt{F}_{\varphi}(x)$ in their deep representation learning, where $\texttt{F}_{\varphi}$ is a deep neural network parameterized by $\varphi$. Specifically, the global MI objective is defined as follows: 
\begin{equation}\label{Eq_JSDMI_G_L}
\small
\begin{split}
\textup{L}_{\theta,\varphi}^{global}(x,y)=\hat{I}_{\theta}^{(JSD)}(x,y)=\hat{I}_{\theta}^{(JSD)}(x,\texttt{F}_{\varphi}(x)).
\end{split}
\end{equation}
The local MI objective aims to maximize the average MI between the high-level representation and feature of local image patches, which intuitively favours information that is shared across patches. Suppose $y=\texttt{F}_{\varphi}(x)=\texttt{f}_{\varphi}\circ  C_{\varphi}(x)$, where $\circ$ denotes the composition operation, $C_{\varphi}(x)=\{C_{\varphi}^{(i)}\}$ encodes the input to a feature map that spatially indexed by $i$, and $\texttt{f}_{\varphi}$ is a subnetwork of $\texttt{F}_{\varphi}$, then the local MI objective is defined as follows:
\begin{equation}\label{Eq_JSDMI_G_L}
\small
\begin{split}
\textup{L}_{{\theta}',\varphi}^{local}(x,y)=\sum_{i}\hat{I}_{{\theta}'}^{(JSD)}(C^{(i)}_{\varphi}(x),y).
\end{split}
\end{equation}

Since identity-related and identity-unrelated information is entangled in the feature representation, the latter could be a discriminate cue for classification and regression during training, thus degrading the generalization ability of the model. We propose to learn to disentangle the identity-related features from the identity-unrelated ones in an unsupervised manner using the above MI estimators. It's worth noting that in view of the loss of information, related for instance to noise, we compute the MI between the output feature of the backbone and the disentangled representation in evaluating the global MI, and the former is also used as the feature map encoded by $C_{\psi}$ in evaluating the local MI. We describe the losses involved in our method in the following.

\noindent\textbf{Global and Local MI Losses. }
The disentangled representation (DR) module in our method consists of two encoders, namely, $E_1$ and $E_2$, splitting the backbone feature $\textbf{f}(Z)=\phi(Z)$ into two parts, i.e., identity-unrelated features $\textbf{f}_u(Z) =  E_1(\textbf{f}(Z))$ and identity-related features $\textbf{f}_r(Z) =  E_2(\textbf{f}(Z))$. These two disentangled features are expected to fully describe $\textbf{f}$, so we maximize the local and global MI between $\textbf{f}$ and the concatenated representation $\tilde{\textbf{f}}(Z)=[\textbf{f}_r(Z),\textbf{f}_u(Z)]$ in the process of learning disentangled representation, resulting in the total loss for MI maximization as follows:
\begin{equation}\label{Eq_MI_loss}
\small
L_{MI}=\rho \textup{L}_{\theta,\varphi}^{global}(Z,\tilde{\textbf{f}}(Z))+ \gamma \textup{L}_{{\theta}',\varphi}^{local}(Z,\tilde{\textbf{f}}(Z)),
\end{equation}
where $\rho$ and $\gamma$ are two constant coefficients to balance the global and local MI terms, respectively, with other losses.

\noindent\textbf{Identity Similarity Loss. }
Suppose the templates $Z$ and $Z'$ have the same identity, i.e., representing one target at different frames. Then the two corresponding identity-related features $E_2(\phi(Z))$ and $E_2(\phi(Z'))$ should be as close as possible for the purpose of classification. So we use the MSE loss to define the identity similarity loss to punish the differences between the two identity-related features, which is formulated by
\begin{equation}
\small
    L_{Idsim}=\omega\left\| E_2(\phi(Z))-E_2(\phi(Z')) \right\|^2_2,
\end{equation}
where $\omega$ is a constant weighted coefficient.

\noindent\textbf{Classification, Regression and Centerness Loss. }
Following P-SiamFC++ \cite{Wang2022RankBasedFP}, the loss for learning the classification and regression tasks is as follows:
\begin{equation}\label{EQ_RRCF}
	\small
	\begin{split}
		L_{CR}= \frac{1}{N_{pos}}\sum_{z}( L_{cls}(p_z, p_z^*) + \lambda_1 I_{\{p_z^*>0\}}L_{qual}(q_z,q_z^*)+\\ \lambda_2 I_{\{p_z^*>0\}}L_{reg}(t_z,t_z^*)),
	\end{split}
\end{equation}
where $z$ represents a coordinate on a feature map, $p_z$is a prediction while $p_z^*$ is the corresponding target label, $I_{\{\cdot\}}$ is the indicator function, $L_{cls}$, $L_{qual}$, and $L_{reg}$ denote the focal loss, the binary cross entropy loss, and the IoU loss for classification, quality assessment, and regression, respectively. Refer to \cite{Wang2022RankBasedFP} for more details. $\lambda_1$ and $\lambda_2$ are weight terms to balance the losses, and $N_{pos}=\sum_{z}I_{\{p_z^*>0\}}$. Note that if $z$ is considered as a positive sample $p_z^*$ is assigned 1, otherwise 0 if considered as a negative sample.

Taken together, the overall loss for training DR-MIM is:
\begin{equation}\label{EQ_totalloss}
\small
	\begin{split}
		L = L_{CR} + L_{MI} + L_{Idsim}.
	\end{split}
\end{equation}

\section{Experiments}
\begin{figure*}[t]
	\centering
	\subfigure{
		\begin{minipage}[t]{0.23\textwidth}
			\includegraphics[width=1\textwidth,height=0.65\textwidth]{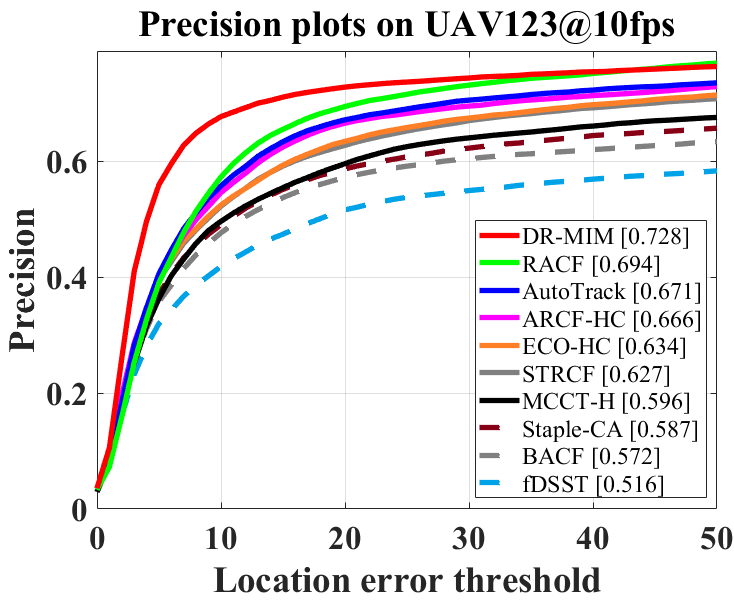}\hspace{0in}
			\includegraphics[width=1\textwidth,height=0.65\textwidth]{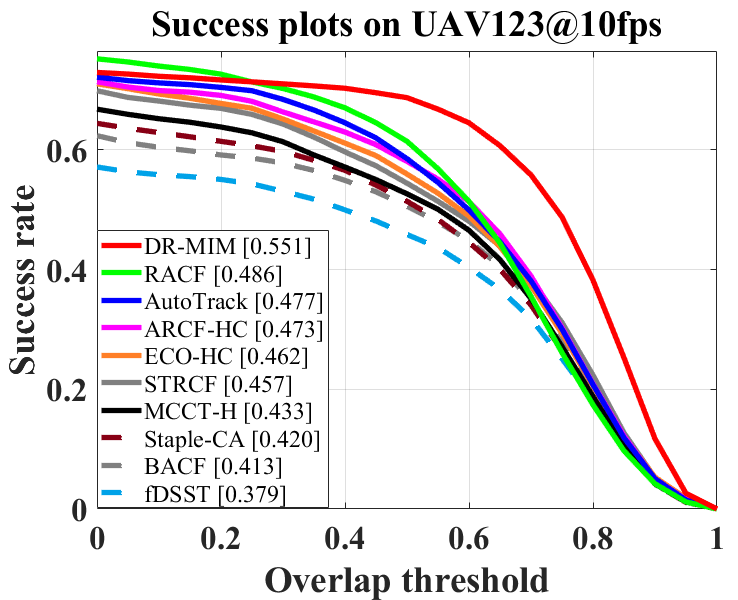}
	\end{minipage}}
	\subfigure{
		\begin{minipage}[t]{0.23\textwidth}
			\includegraphics[width=1\textwidth,height=0.65\textwidth]{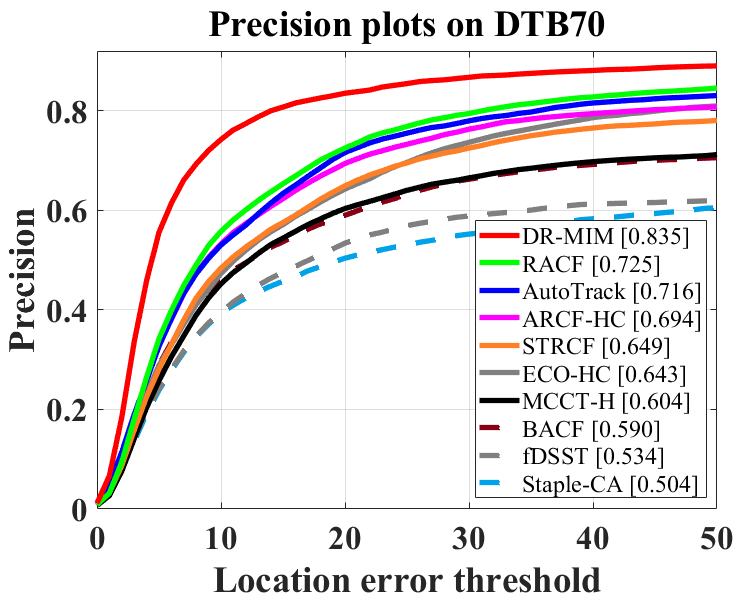}\hspace{0in}
			\includegraphics[width=1\textwidth,height=0.65\textwidth]{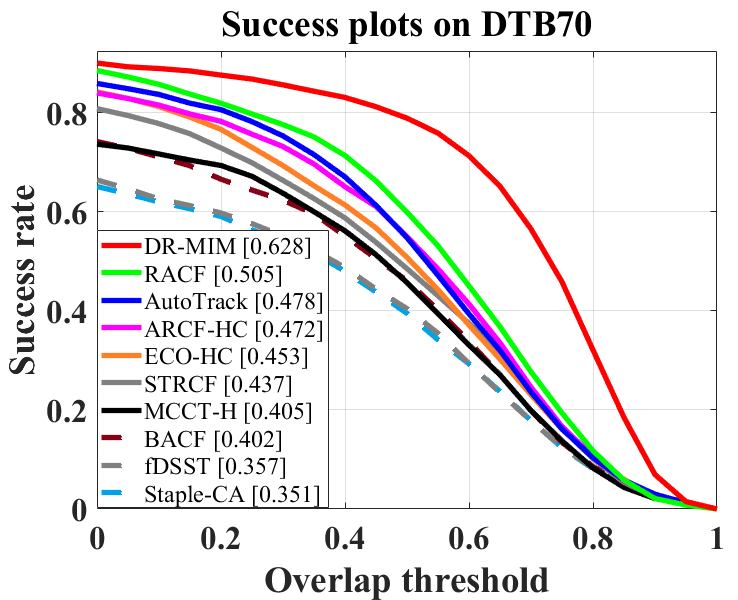}
	\end{minipage}}
	\subfigure{
		\begin{minipage}[t]{0.23\textwidth}
			\includegraphics[width=1\textwidth,height=0.65\textwidth]{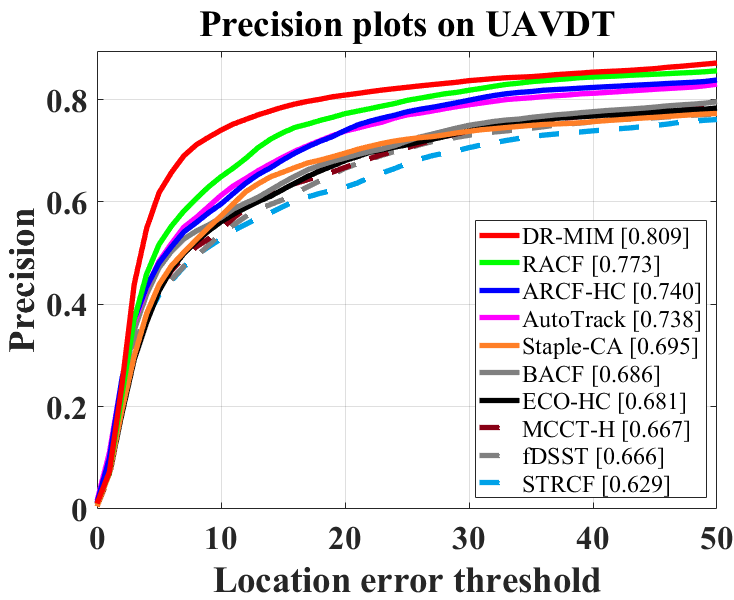}\hspace{0in}
			\includegraphics[width=1\textwidth,height=0.65\textwidth]{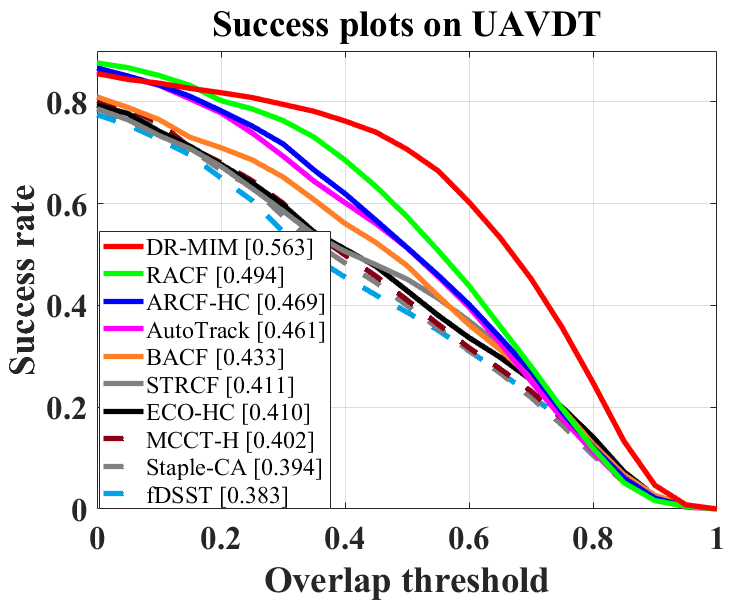}
	\end{minipage}}
	\subfigure{
		\begin{minipage}[t]{0.23\textwidth}
			\includegraphics[width=1\textwidth,height=0.65\textwidth]{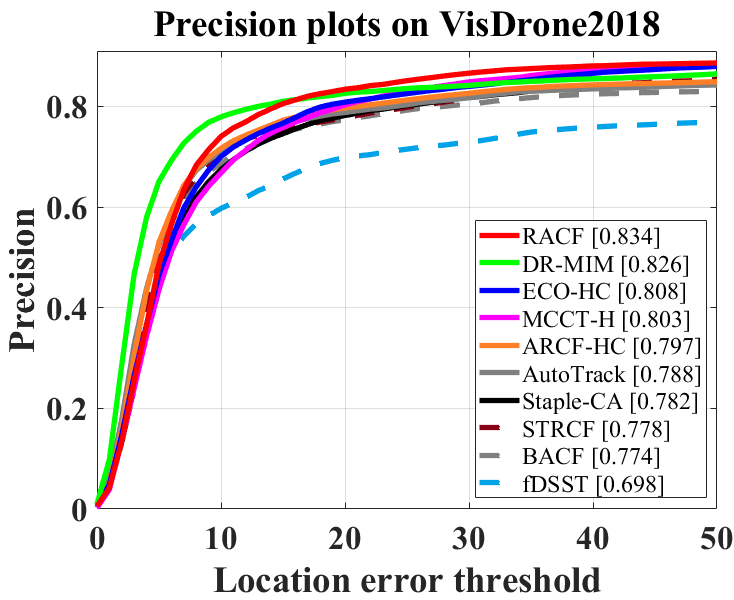}\hspace{0in}
			\includegraphics[width=1\textwidth,height=0.65\textwidth]{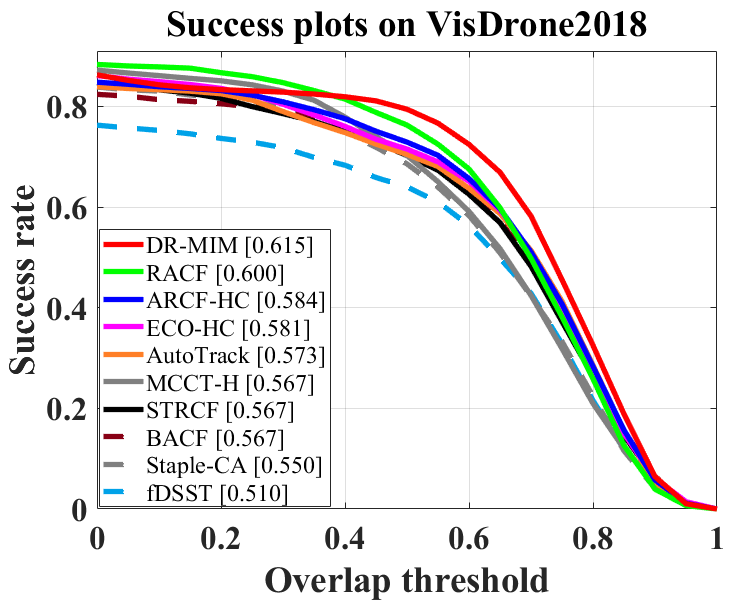}
	\end{minipage}}
	\caption{Overall performance of hand-crafted based trackers on UAV123@10fps, DTB70, UAVDT, and VisDrone2018.}
	\label{fig_overall_p_s_plots}
\end{figure*}

\begin{table*}
				\centering
				\begin{minipage}[h]{\textwidth}
				\caption{Average precision and speed (FPS) comparision between DR-MIM and hand-crafted based trackers on UAV123@10fps, DTB70, UAVDT and VisDrone2018.  All the reported FPSs are evaluated on a single CPU. {\color[HTML]{FE0000}Red}, {\color[HTML]{3531FF}blue} and {\color[HTML]{009901}green} respectively indicate the first, second and third place.}
	\label{tab:precision_FPS}
	\resizebox{7.0in}{0.32in}{
		\begin{tabular}{@{}cccccccccccccc    ccc@{}}
			\toprule
				& KCF\cite{2015High}& fDSST\cite{danelljan2016adaptive}& BACF\cite{2017Learning}& ECO-HC\cite{danelljan2017eco}&  STRCF\cite{li2018learning} & ARCF-HC\cite{huang2019learning}  & AutoTrack\cite{li2020autotrack} 
		   &RACF\cite{li2021learning} & F-SiamFC++ \cite{Wu2022FisherPF}  & P-SiamFC++\cite{Wang2022RankBasedFP} & \textbf{DR-MIM (Ours)}                                   \\ \hline\hline
			\textbf{Precision} 
			& 53.3& 60.4& 64.2  & 68.8  	 & 67.1   & 71.9 	&72.3 
				 &  75.7 &\color[HTML]{009901}\textbf{78.4}	& {\color[HTML]{3531FF} \textbf{78.8}}  & {\color[HTML]{FE0000} \textbf{80.2}} 
			\\
			\textbf{FPS (CPU)} & {\color[HTML]{FE0000} \textbf{622.5}}&  {\color[HTML]{3531FF} \textbf{193.4}}& 54.2  
   & 84.5 & 28.4 	 & 34.2     & 58.7    & 35.7  &51.9 & 46.1   &  {\color[HTML]{009901} \textbf{92.1}}  
   \\ \hline
		\end{tabular}
	}
				\end{minipage}
				\begin{minipage}[h]{\textwidth}
				\caption{Precision and speed (FPS) comparison between DR-MIM and
		deep-based trackers on VisDrone2018 \cite{du2018the}. All the reported FPSs are evaluated on a single GPU. {\color[HTML]{FE0000}Red}, {\color[HTML]{3531FF}blue} and {\color[HTML]{009901}green} indicate the first, second and third place. }
	\label{tab:precision_FPS_deep}
 	\resizebox{7.0in}{0.32in}{
		\begin{tabular}{cccccccccc}
			\toprule
   &\begin{tabular}[c]{@{}c@{}}SiamGAT\cite{guo2021graph}\end{tabular} 
   &  \begin{tabular}[c]{@{}c@{}}HiFT\cite{cao2021hift}\end{tabular}
   & \begin{tabular}[c]{@{}c@{}}AutoMatch\cite{zhang2021learn}\end{tabular} 
   
         &\begin{tabular}[c]{@{}c@{}}SLT-SiamRPN++\cite{Kim2022TowardsST}\end{tabular} 
        
         &\begin{tabular}[c]{@{}c@{}}TCTrack\cite{Cao2022TCTrackTC}\end{tabular}
         &\begin{tabular}[c]{@{}c@{}}F-SiamFC++\cite{Wu2022FisherPF}\end{tabular}
         &\begin{tabular}[c]{@{}c@{}}P-SiamFC++\cite{Wang2022RankBasedFP}\end{tabular}
          &\begin{tabular}[c]{@{}c@{}}SparseTT\cite{2022SparseTT}\end{tabular}
     & \begin{tabular}[c]{@{}c@{}}\textbf{DR-MIM (Ours)} \end{tabular} 
          \\ \hline\hline
          
	\textbf{Precision}  &78.3 &71.9    &78.1   & 79.4          &78.4 & 80.7                                        & {\color[HTML]{009901} \textbf{80.9}}                   & {\color[HTML]{3531FF} \textbf{81.4}} 
 & {\color[HTML]{FE0000} \textbf{82.6}}  
 
    \\\textbf{FPS (GPU)}       
 
			 &75.6 &132.7  &47.9 &32.2  &132.2  &{\color[HTML]{3531FF} \textbf{251.5}}  &{\color[HTML]{009901} \textbf{230.5}} &42.2         & {\color[HTML]{FE0000} \textbf{363.9}}    \\ \hline
		\end{tabular}
	}
				\end{minipage}
\end{table*}

\begin{table*}
\centering
\caption{Comparison of model size (parameters), precision and tracking speed between proposed DR-MIM and the baseline method P-SiamFC++ on four UAV benchmarks. PRC is short for precision. Note that only the precision on CPU is shown here since the difference of precision on CPU and GPU is very small.}
\label{tab:compareWbaseline}
\resizebox{7.0in}{0.55in}{
\begin{tabular}{ccccccccccccccccc} 
\toprule
\multirow{3}{*}{Methods} & \multirow{3}{*}{Parameters} & \multicolumn{3}{c}{UAV123@10fps}                      & \multicolumn{3}{c}{DTB70}                             & \multicolumn{3}{c}{UAVDT}                             & \multicolumn{3}{c}{VisDrone2018}                      & \multicolumn{3}{c}{Avg.}                               \\ 
\cmidrule(l){3-17}
                         &                             & \multirow{2}{*}{PRC} & \multicolumn{2}{c}{FPS}        & \multirow{2}{*}{PRC} & \multicolumn{2}{c}{FPS}        & \multirow{2}{*}{PRC} & \multicolumn{2}{c}{FPS}        & \multirow{2}{*}{PRC} & \multicolumn{2}{c}{FPS}        & \multirow{2}{*}{PRC} & \multicolumn{2}{c}{FPS}         \\ 
\cmidrule(l){4-5}\cmidrule(l){7-8}\cmidrule(l){10-11}\cmidrule(l){13-14}\cmidrule(l){16-17}
                         &                             &                      & CPU           & GPU            &                      & CPU           & GPU            &                      & CPU           & GPU            &                      & CPU           & GPU            &                      & CPU           & GPU             \\ 
\hline\hline
P-SiamFC++ \cite{Wang2022RankBasedFP}              & 7.49M                       & \textbf{73.1}        & 45.1          & 236.4          & 80.3                 & 45.6          & 238.2          & 80.7                 & 48.8          & 258.8          & 80.9                 & 45.0          & 230.5          & 78.8                 & 46.1          & 241.0           \\
\textbf{DR-MIM (Ours)}            & \textbf{4.83M}                       & 72.8                 & \textbf{91.7} & \textbf{370.2} & \textbf{83.5}        & \textbf{89.1} & \textbf{377.9} & \textbf{80.9}        & \textbf{94.5} & \textbf{370.6} & \textbf{82.6}        & \textbf{93.1} & \textbf{363.9} & \textbf{80.2}        & \textbf{92.1} & \textbf{370.7}  \\
\bottomrule
\end{tabular}
}
\end{table*}

We conduct our experiments on four challenging UAV benchmarks, i.e., UAV123@10fps \cite{2016A}, DTB70 \cite{li2017visual}, UAVDT \cite{du2018the} and VisDrone2018 \cite{wen2018visdrone}. All evaluation experiments are conducted on a PC equipped with i9-10850K processor (3.6GHz), 16GB RAM and an NVIDIA TitanX GPU. 
All the block-wise pruning ratios for the backbone (AlexNet), the neck, and the head are 0.5. The constants $\rho$, $\gamma$, and $
\omega$ are all set to 0.05.
Other parameters for training and inference follow P-SiamFC++ \cite{Wang2022RankBasedFP}. 
Code will be available on: https://github.com/P-SiamFCpp/DR-MIM.

\subsection{Comparison with CPU-based Trackers}  
Eight state-of-the-art trackers based on hand-crafted features for comparison are: KCF \cite{2015High}, fDSST \cite{danelljan2016adaptive},  BACF \cite{2017Learning}, ECO-HC \cite{danelljan2017eco}, STRCF \cite{li2018learning}, ARCF-HC \cite{huang2019learning}, AutoTrack  \cite{li2020autotrack}, RACF \cite{li2021learning}.  
The overall performance of DR-MIM with the competing trackers on the four benchmarks is shown in Fig. \ref{fig_overall_p_s_plots}. It can be seen that DR-MIM outperforms all other trackers on all four benchmarks except for the VisDrone2018.
Specifically, on UAV123@10fps, DTB70 and UAVDT, DR-MIM significantly outperforms the second tracker RACF in terms of precision and AUC (i.e., area under curve), with gains of (3.4\%, 6.5\%), (11.0\%, 12.3\%) and (3.6\%, 6.9\%), respectively. On VisDrone2018, DR-MIM is inferior to the first tracker RACF in precision with a small gap of 0.8\%. However, we rank first in terms of AUC. Note that the parameters of RACF are dataset-specific. In terms of speed, we use the average FPS over the aforementioned four benchmarks on CPU as the metric. Table \ref{tab:precision_FPS} illustrates average precision and FPS produced by different trackers. As can be seen, DR-MIM outperforms all the competing trackers in precision and is also the best real-time tracker (speed of $>$30FPS) on CPU. Specifically, DR-MIM achieves 80.2\% in precision at a speed of 92.1 FPS.

\subsection{Comparison with Deep Learning-based Trackers}
The proposed DR-MIM is also compared with eight state-of-the-art
deep trackers on the VisDrone2018 dataset \cite{wen2018visdrone}, including SiamGAT \cite{guo2021graph}, HiFT \cite{cao2021hift}, AutoMatch \cite{zhang2021learn}, SLT-SiamRPN++ \cite{Kim2022TowardsST}, SparseTT \cite{2022SparseTT},
TCTrack \cite{Cao2022TCTrackTC}, F-SiamFC++ \cite{Wu2022FisherPF}, P-SiamFC++ \cite{Wang2022RankBasedFP}.

The FPSs and the precisions on VisDrone2018 are shown in Table \ref{tab:precision_FPS_deep}. As can be seen, the precision and the GPU speed of our DR-MIM outperform that of the competing DL-based trackers, and its GPU speed is 8 times faster than the second tracker SparseTT \cite{2022SparseTT}. This further justifies the effectiveness of the proposed method of learning disentangled representation with MI maximization for real-time UAV tracking. 

\subsection{Qualitative Evaluation}
\begin{figure}[h]
	\centering
	\includegraphics[width=0.475\textwidth,height=0.3\textwidth]{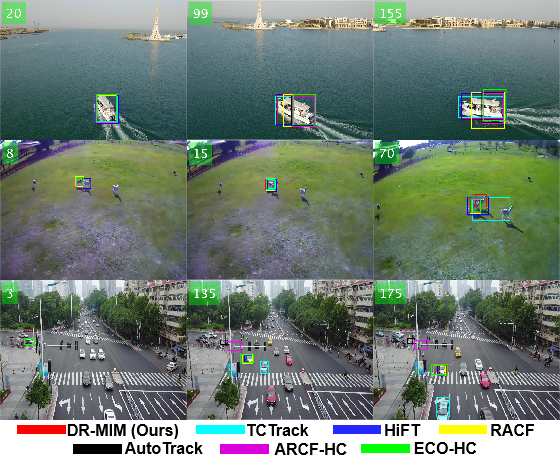}
	\caption{Qualitative evaluation on 3 sequences from, respectively, UAV123@10fps, DTB70 and VisDrone2018 (i.e. boat3, Animal3 and uav0000294\_00000\_s). } \label{fig:visual_examples}
\end{figure}

We show some qualitative tracking results of our method and six state-of-the-art trackers in Fig. \ref{fig:visual_examples}. It can be seen that other trackers fail to maintain robustness in challenging examples where objects are experiencing background cluster (i.e., Animal3) or pose change (i.e., boat3 and uav0000294\_00000\_s), but our DR-MIM performs much better and is visually more satisfying. This qualitatively shows learning disentangled representation with MIM can improve tracking precision.

\subsection{Ablation Study}
\indent\textbf{Effect of disentangled representation:} 
We compare the proposed DR-MIM with the baseline P-SiamFC++ on all four UAV benchmarks in terms of model size, precision, and tracking speed to understand its effectiveness. Their comparisons are shown in Table \ref{tab:compareWbaseline}. As can be seen, the model size of DR-MIM is reduced to 64.5\% ($\approx$4.83/7.49) of the original. Both the CPU and GPU speeds are improved. Since the parallel computing units on our GPU far exceed the size of both models, the GPU speed increases by only 53.8\% on average. But the average CPU speed is raised from 46.1 FPS to 92.1 FPS, almost two times faster. Although DR-MIM is slightly inferior to the baseline on UAV123@10fps with a gap of 0.3\% in precision, the improvement on DTB70 and VisDrone2018 is apparent, specifically, with gains of 3.2\% and 1.7\%, respectively. These results justify that the proposed method is effective in improving both efficiency and precision.

\noindent\textbf{Impact of the global pruning ratio:} To see how the global
pruning ratio $\mu$ affects the precision of DR-MIM, we trained DR-MIM with different pruning ratios. Specifically, each convolutional layer in the backbone, neck and head is pruned with the same global ratio, which ranges from 0.1 to 0.8. 
The precisions of DR-MIM on DTB70 with different pruning ratios are shown in Table \ref{tab:Impact_of_ratios}. As can be seen, the best precision happens when the pruning ratio is 0.5, which is also the default setting of our DR-MIM. These results suggest that with the proposed disentangled representation learning we are able to boost both precision and efficiency for UAV tracking if the global pruning ratio is set appropriately.

\begin{table}
\centering
\caption{Illustration of how the precision of DR-MIM on DTB70 varies with the global pruning ratio $\mu$. }
\label{tab:Impact_of_ratios}
\resizebox{3.35in}{0.25in}{
\begin{tabular}{ccccccccc} 
\toprule
$\mu$ & 0.1  & 0.2                             & 0.3  & 0.4                                       & 0.5                            & 0.6  & 0.7  & 0.8   \\ \hline
\midrule
\textbf{Precision}  & 78.7 & \textbf{\textcolor{blue}{82.5}} & 77.1 & \textbf{\textcolor[rgb]{0,0.502,0}{80.3}} & \textbf{\textcolor{red}{83.5}} & 77.6 & 76.3 & 71.2  \\
\bottomrule
\end{tabular}
}
\end{table}

\section{Conclusion}

In this work, we are the first to explore learning disentangled representation via mutual information maximization for UAV tracking. The proposed DR-MIM is able to learn more effective and more compact representations, and demonstrates state-of-the-art performance on four UAV benchmarks. The proposed method can not only improve efficiency but, surprisingly, also improve tracking precision. We believe our work will draw more attention to developing lightweight DL-based trackers for UAV tracking.

\section*{Acknowledgment}
Thanks to the supports by Guangxi Key Laboratory of Embedded Technology and Intelligent System and the Guangxi Science and Technology Base and Talent Special Project (No. 2021AC9330).

\small
\bibliographystyle{IEEEbib}
\bibliography{icme}

\begin{thebibliography}{10}

\bibitem{li2020autotrack}
Li~Y. and et~al.,
\newblock ``Autotrack: Towards high-performance visual tracking for uav with
  automatic spatio-temporal regularization,'' 2020,CVPR,01194.

\bibitem{cao2021hift}
Cao Z. and et~al.,
\newblock ``Hift: Hierarchical feature transformer for aerial tracking,''
\newblock in {\em ICCV}, 2021, pp. 15457--15466.

\bibitem{Wang2022RankBasedFP}
Wang X. and et~al.,
\newblock ``Rank-based filter pruning for real-time uav tracking,''
\newblock in {\em ICME}, 2022, pp. 01--06.

\bibitem{Wu2022FisherPF}
Wu~W., P.~Zhong, and et~al.,
\newblock ``Fisher pruning for real-time uav tracking,''
\newblock in {\em IJCNN}, 2022, pp. 1--7.

\bibitem{articlewxc}
Wang X. and et~al.,
\newblock ``Exploiting rank-based filter pruning for real-time uav tracking,''
\newblock {\em SSRN Electronic Journal}, 01 2022.

\bibitem{li2021learning}
Li~S. and et~al.,
\newblock ``Learning residue-aware correlation filters and refining scale for
  real-time uav tracking,''
\newblock {\em PR}, vol. 127, pp. 108614, 2022.

\bibitem{huang2019learning}
Z.~H and et~al.,
\newblock ``Learning aberrance repressed correlation filters for real-time uav
  tracking,''
\newblock in {\em ICCV,2019}, pp. 2891--2900.

\bibitem{Zhang2022TrackingSA}
Zhang Z. and et~al.,
\newblock ``Tracking small and fast moving objects: A benchmark,''
\newblock in {\em ACCV}, 2022.

\bibitem{Zhang2023TsfmoAB}
Zhang Z. and et~al.,
\newblock ``Tsfmo: A benchmark for tracking small and fast moving objects,''
\newblock {\em SSRN Electronic Journal}, 2023.

\bibitem{Liu2022GlobalFP}
Liu M. and et~al.,
\newblock ``Global filter pruning with self-attention for real-time uav
  tracking,''
\newblock in {\em BMVC}, 2022.

\bibitem{Zhong2022EfficiencyAP}
Zhong P. and et~al.,
\newblock ``Efficiency and precision trade-offs in uav tracking with filter
  pruning and dynamic channel weighting,''
\newblock in {\em FSDM}, 2022.

\bibitem{Zhang2020FaceAV}
Zhang K. and et~al.,
\newblock ``Face anti-spoofing via disentangled representation learning,''
\newblock {\em ArXiv}, vol. 08250, 2020.

\bibitem{Yuan2021ImprovingZV}
Yuan S. and et~al.,
\newblock ``Improving zero-shot voice style transfer via disentangled
  representation learning,''
\newblock {\em ArXiv}, vol. 09420, 2021.

\bibitem{Zhao2021LearningVH}
Zhao L. and et~al.,
\newblock ``Learning view-disentangled human pose representation by contrastive
  cross-view mutual information maximization,''
\newblock {\em CVPR}, pp. 12788--12797, 2021.

\bibitem{Guo2019LearningDR}
Guo W. and et~al.,
\newblock ``Learning disentangled representation for cross-modal retrieval with
  deep mutual information estimation,''
\newblock {\em 27th ACM ICM}, 2019.

\bibitem{tschannen2019mutual}
Michael T. and et~al.,
\newblock ``On mutual information maximization for representation learning,''
\newblock in {\em ICLR}, 2019.

\bibitem{Li2021LearningRC}
Li~S. and et~al.,
\newblock ``Learning residue-aware correlation filters and refining scale
  estimates with the grabcut for real-time uav tracking,''
\newblock {\em 3DV}, pp. 1238--1248, 2021.

\bibitem{li2020asymmetric}
Li~S. and et~al.,
\newblock ``Asymmetric discriminative correlation filters for visual
  tracking,''
\newblock {\em FITEE}, vol. 21, no. 10, pp. 1467--1484, 2020.

\bibitem{Li2021EquivalenceOC}
Li~S. and et~al.,
\newblock ``Equivalence of correlation filter and convolution filter in visual
  tracking,''
\newblock {\em ArXiv}, vol. abs/2105.00158, 2021.

\bibitem{Cao2022TCTrackTC}
Cao Z. and et~al.,
\newblock ``Tctrack: Temporal contexts for aerial tracking,''
\newblock {\em CVPR}, pp. 14778--14788, 2022.

\bibitem{Linsker1988SelforganizationIA}
Linsker R.,
\newblock ``Self-organization in a perceptual network,''
\newblock {\em Computer}, vol. 21, pp. 105--117, 1988.

\bibitem{Li2022TowardsLD}
Li~Y. and et~al.,
\newblock ``Towards learning disentangled representations for time series,''
\newblock {\em ACM SIGKDD}, 2022.

\bibitem{Poole2019OnVB}
Poole B. and et~al.,
\newblock ``On variational bounds of mutual information,''
\newblock in {\em ICML}, 2019.

\bibitem{Hjelm2019LearningDR}
Hjelm R.D. and et~al.,
\newblock ``Learning deep representations by mutual information estimation and
  maximization,''
\newblock {\em ArXiv}, vol. abs/1808.06670, 2019.

\bibitem{2015High}
Henriques J.F. and et~al.,
\newblock ``High-speed tracking with kernelized correlation filters,''
\newblock {\em IEEE TPAMI,2015}, vol. 37, no. 3, pp. 583--596.

\bibitem{danelljan2016adaptive}
Danelljan M. and et~al.,
\newblock ``Adaptive decontamination of the training set: A unified formulation
  for discriminative visual tracking,''
\newblock in {\em CVPR,2016}, pp. 1430--1438.

\bibitem{2017Learning}
Hamed G., F.~Ashton, and et~al.,
\newblock ``Learning background-aware correlation filters for visual
  tracking,''
\newblock in {\em ICCV}, 2017.

\bibitem{danelljan2017eco}
Martin D. and et~al.,
\newblock ``Eco: Efficient convolution operators for tracking,''
\newblock in {\em CVPR,2017}, pp. 6931--6939.

\bibitem{li2018learning}
Li~F. and et~al.,
\newblock ``Learning spatial-temporal regularized correlation filters for
  visual tracking,''
\newblock in {\em CVPR,2018}, pp. 4904--4913.

\bibitem{du2018the}
Du~D. and et~al.,
\newblock ``The unmanned aerial vehicle benchmark: Object detection and
  tracking,''
\newblock in {\em ECCV,2018}, pp. 375--391.

\bibitem{guo2021graph}
Guo D. and et~al.,
\newblock ``Graph attention tracking,''
\newblock in {\em CVPR,2021}, pp. 9543--9552.

\bibitem{zhang2021learn}
Zhang Z. and et~al.,
\newblock ``Learn to match: Automatic matching network design for visual
  tracking,''
\newblock in {\em ICCV,2021}, pp. 13339--13348.

\bibitem{Kim2022TowardsST}
Kim M. and et~al.,
\newblock ``Towards sequence-level training for visual tracking,''
\newblock {\em ArXiv}, vol. abs/2208.05810, 2022.

\bibitem{2022SparseTT}
Fu~Z. and et~al.,
\newblock ``Sparsett: Visual tracking with sparse transformers,''
\newblock {\em ArXiv}, vol. abs/2205.03776, 2022.

\bibitem{2016A}
Matthias M. and et~al.,
\newblock ``A benchmark and simulator for uav tracking,''
\newblock {\em FJMS,2016}, vol. 2, no. 2, pp. 445--461.

\bibitem{li2017visual}
Li~S. and et~al.,
\newblock ``Visual object tracking for unmanned aerial vehicles: A benchmark
  and new motion models.,''
\newblock in {\em AAAI,2017}, pp. 4140--4146.

\bibitem{wen2018visdrone}
Wen L. and et~al.,
\newblock ``Visdrone-sot2018: The vision meets drone single-object tracking
  challenge results,''
\newblock in {\em ECCV}, 2018, pp. 469--495.

\end{thebibliography}

\end{document}